\documentclass[conference,compsoc]{IEEEtran}
\usepackage{amsmath}
\usepackage{amssymb}
\usepackage{textcomp}
\usepackage{tikz}
\usepackage{longtable}
\usepackage{adjustbox}
\usepackage{float}
\floatstyle{boxed} 
\usepackage{graphicx}
\usepackage[noend]{algpseudocode}
\usepackage{algorithm}
\usepackage{booktabs}

\usetikzlibrary{positioning}

\ifCLASSOPTIONcompsoc
  \usepackage[nocompress]{cite}
\else
  \usepackage{cite}
\fi
\ifCLASSINFOpdf
\else
\fi
\definecolor{darkgray}{rgb}{0.37,0.37,0.52}
\definecolor{darkgreen}{rgb}{0.0,0.55,0.0}

\newcommand{\prosaic}[1]{#1}

\begin{document}
%\title{Binarized Convolutional Neural Networks for Efficient Real-Time Inference [On GPU]}
\title{Binarized Convolutional Neural Networks for Efficient Inference on GPUs}

%\author{\IEEEauthorblockN{Mir Khan, Heikki Huttunen, Olli Suominen and Atanas Gotchev}
%\IEEEauthorblockA{Laboratory of Signal Processing\\
%Tampere University of Technology\\
%Tampere, Finland\\
%Email: mir.khan@tut.fi, heikki.huttunen@tut.fi, olli.j.suominen@tut.fi, atanas.gotchev@tut.fi}}

\author{\IEEEauthorblockN{Mir Khan, Heikki Huttunen, Jani Boutellier}
\IEEEauthorblockA{%Laboratory of \textbf{LABORATORY\_NAME}\\
Tampere University of Technology\\
Tampere, Finland\\
Email: Mir.Khan@tut.fi, Heikki.Huttunen@tut.fi, Jani.Boutellier@tut.fi}}

\maketitle

\begin{abstract}
%\jani{usually the abstract starts with a sentence or two that clarify the general contect of the work. Here that would be CNNs for embedded platforms}

Convolutional neural networks have recently achieved significant breakthroughs in various image classification tasks. However, they are computationally expensive, which can make their feasible implementation on embedded and low-power devices difficult. In this paper convolutional neural network binarization is implemented on GPU-based platforms for real-time inference on resource constrained devices. In binarized networks, all weights and intermediate computations between layers are quantized to +1 and -1, allowing multiplications and additions to be replaced with bit-wise operations between 32-bit words. This representation completely eliminates the need for floating point multiplications and additions and decreases both the computational load and the memory footprint compared to a full-precision network implemented in floating point, making it well-suited for resource-constrained environments. %Our approach is demonstrated with a convolutional neural network that is used for vehicle type classification. 
We compare the performance of our implementation with an equivalent floating point implementation on one desktop and two embedded GPU platforms. Our implementation achieves a maximum speed up of $\mathbf{7.4}\times$ with only 4.4\% loss in accuracy compared to a reference implementation.

%[we verify and show that the binary net method works on car data and it's effective. We implement a binarized convolutional neural network for vehicle classification. We [also] study the impact of input binarization on speed and accuracy using various input binarization methods such as  automatic-preprocessing conv layer,  thresholding..]
\end{abstract}

\makeatletter
\def\ps@IEEEtitlepagestyle{
  \def\@oddfoot{\mycopyrightnotice}
  \def\@evenfoot{}
}
\def\mycopyrightnotice{
  {\footnotesize
  \begin{minipage}{\textwidth}
  \centering
  Copyright~\copyright~2018 IEEE. Accepted to Proc. IEEE EUSIPCO 2018, Sep. 3-7, 2018, Rome, Italy
  \end{minipage}
  }
}
%\response{I think it's pretty much done, apart from the other green comments. So if there are any suggestions it should be done today or tomorrow in the morning. Deadline is 2 PM (I think) tomorrow, but I'll submit a copy today at the end of the day.}

% \fi
\IEEEpeerreviewmaketitle

\vspace{-0.75mm}
\section{Introduction}
\vspace{-0.75mm}
%\textit{(This section + previous work, + abstract will take more time...)}\\


%[Some introduction about neural nets]...
In the recent years, convolutional neural networks (CNNs) have presented impressive performance in image classification [\ref{imageclassification}][\ref{augmentimage1}], face recognition [\ref{facerecognition1}][\ref{facerecognition2}], audio classification [\ref{audioclassification2}], and speech recognition [\ref{speechrecognition}]."

%\jani{the section below is not mandatory, but if you have space you can include a CNN introduction like that}

%\expendable {2} \prosaic{ Convolutional neural etworks are like standard nets but Typically it has 1 or more convolutional layers . they are well suited for extracting information from images and they've been shown good results (refs). the conv layer outputs feature maps extracted from the data by filtering the image with a number of kernels. following the conv layer is usually 1 ir nmore a standard Describe how convolution is done as matmult (ref conv2gemm)}

Large neural network models can be computationally expensive, making them unsuitable for deployment to small resource-constrained mobile devices. To this extent, contemporary CNN-based solutions often acquire the input data on a mobile device, but transmit the data to a remote server for CNN-based processing. However, performing the CNN-based processing on the mobile device (a.k.a. edge computing) would reduce the overall system complexity and enable real-time applications. %\curtail{such as those related to Augmented Reality []}.

The emerging CNN subfield of \textit{model compression} aims to retain the accuracy of the neural network while minimizing redundant network parameters and reducing computational load. Many such techniques have already been proposed.

%\jani{Please check whether we should speak about 'parameters' or 'weights' here. I'm not sure they're interchangeable}\response{ : I think they are interchangeable. I also checked the original pruning paper, and they use both words to describe the weights of the network.}

One technique [\ref{refPruning}] is based on \textit{pruning} of parameters, where majority of the parameters of the network are removed without significantly impacting accuracy. Reduction of parameters initially leads to a significant drop in accuracy; however, retraining (fine-tuning) of the parameters restores most of the network's accuracy. The authors report 13$\times$ reduction of memory requirements with no loss in accuracy [\ref{refPruning}].

Another approach, low-rank approximation of convolutional kernels [\ref{refSepConv}], approximates 2D convolutions with convolutions by vectors. The separable kernels can be obtained either by training the network with separable filters [\ref{refDecompose}] or by posing it as an optimization problem to minimize the reconstruction error of the feature maps. Depending on the approach [\ref{refSepConv}][\ref{refDecompose}], speedups between 2$\times$ to 4$\times$ have been reported on CPU implementations.

%\begin{equation}
%argmin_{\{h\} , \{v\}} E= \frac{1}{M}\sum(conv(X,W))\\ - \frac{1}{M}\sum(colconv(rowconv(X,rowkernel),colkernel))
%\end{equation}

Binarized neural networks (BNN) have been first introduced in [\ref{refBinNet}], where their performance was demonstrated on the CIFAR-10 dataset. The weights and activations for intermediate computations are binarized to $+1$ and $-1$. The authors present a speed up of 7$\times$ on a network for the MNIST dataset. In a further work [\ref{refXnor}] the approach was refined for CPU implementation and evaluated on the ImageNet dataset.

by packing 1-bit weights into 32-bit words, enabling replacement of multiplication operations by logic XNORs.
In this paper, an approach for the implementation of
BNNs [11] on GPU platforms is presented. To the best
knowledge of the authors, this is the first work that presents
a GPU implementation of a binarized convolutional neural
network for inference. We present our implementation with
an application use case of vehicle type classification [12].
Results show significant speedups in real-time inference
compared to a floating point version of an equivalent neural
network.

As a summary, the contributions of this work are as follows:
\vspace{-0.8mm}
\begin{itemize}
	\item Detailed presentation of efficiently implementing
CNN binarization, including the convolutional layers,
on GPU-based platforms.
    \item Comparison of different approaches for binarizing
input data, and how each approach impacts the classification
accuracy.
    \item Performance (execution time) comparisons on several
platforms.
    
\end{itemize}

The source code for our CUDA implementation is publically available \footnote{ github.com/Valentin4869/BinCNN}{}.

\section{Experimental Setup}
\subsection{Binarizing the network}
Our binarized network architecture is based on the original vehicle classifier network presented in [\ref{refHeikkiCar}]. We implement a binarized version of the same architecture in several steps. We do not use any ReLU [\ref{refRELU}] activations in the binarized version. In the original binarization work [\ref{refBinNet2}], the authors suggest two approaches for binarization: stochastic and deterministic. For binarizing the weights and intermediate computations, we use the deterministic \textit{sign} function, which is defined as

\begin{equation}
sign(x) = \begin{cases} -1 & \text{if } x \leq 0 \\ +1 & \text{if } x > 0 \end{cases}
\end{equation}
For training the BNN, following [\ref{refSTE}], we explicitly define the gradient of the \textit{sign} function to be the identity function in the backward pass, such that $\frac{\partial sign(x)}{\partial x} = x$.

The non-binarized network is trained with the RMSprop optimizer [\ref{refRMSprop}], while the binarized version is trained with the ADAM [\ref{refAdam}] optimizer.  After training, only the binarized weights are used for inference for the binarized network.

The network is trained with a dataset set consisting of 6555 images of vehicles that have been captured by a camera and manually categorized into four different classes: \textit{bus}, \textit{normal}, \textit{truck}, and \textit{van}. Each image has size $96\times 96$ and are in full color. The data has been split into a training set (90\%) and a test set (10\%).  We augment the training set using flipping and filtering with a 2D Gaussian filter with $\hspace {-0.8mm} \sigma=  0.5$, resulting in a total training set size of 14,108 images, 20\% of which are used for validation. Throughout this text, our accuracy reports are for the performance of the network on the test set that corresponds to the best validation set accuracy.

%\response{Maybe we shouldn't include a net  diagram? Takes too much space and no advantage compared to a written description of the net architecture. Maybe there is no need to describe the architecture in detail since we refer to the original net in heikkiCar and how we modify/binarize it? I could still make the diagram at the end of today if it's needed.}
%\jani{Yes, you can omit the diagram as we are overlength}

\subsection{Testing pipeline}

%option 1: Measurement starts from memcpyHD, kernel launch, and ends at memcpyDH. \\
%option 2: Allocate space for more than one input, Measurement starts from first memcpyHD , kernels launched, then second memcpyHD, launch kernels.... when batch is completed, we wait and do memcpyDHasync from separate buffers.\\
%option 3 : memcpy, starttimer, launch kernels, wait for last kernel to finish, stop timer. Motivation: memcpyHD/DH can vary depending on copy method, platform, whether batch processing is used..., so it is probably fair to only measure computation times for the kernels, and average those for different inputs.

For obtaining runtime results, we use the built-in GPU timers to measure the runtime of the kernels for our CUDA and OpenCL programs. Our kernel execution time measurements do not include memory transfer times to/from the GPU, as they can be affected by various factors, some of which are hardware-dependent, for example, on the NVidia Jetson host and device memory are shared. The correctness and accuracy of the profiling results generated have been verified by the Nvidia Visual Profiler for the same CUDA programs.

For each test run, 1000 images are randomly generated and fed to the network one at a time. The timer begins after the memory is copied, and the timer ends after the last kernel's computation is completed. Our final result is the total accumulated time per sample averaged over all 1000 samples.

\subsection{Input binarization}
In this section we describe our methods for binarizing the inputs to the first layer of our BNN. We pre-process the data set using these techniques and evaluate the accuracy of the BNN on the pre-processed data set.

\textbf{Thresholding}
A constant threshold $T$ can be subtracted from the input $\mathbf{X}$ before binarizing it. We simply substitute the input $\bf{X}$ to the first layer with $sign(\mathbf{X} + T)$, for $\mathbf{X}\in \mathbb{R}^{H\times W\times C}$, and for $T \in \mathbb{R}^{1\times 1\times C}$. The motivation is to shift the range of values taken by $\mathbf{X}$ such that binarization with the \textit{sign} function produces meaningful results, as opposed to all zeros for standard pixel-value ranges do not include negative numbers. The network is trained as before but in two stages: first, the network is trained for $50$ epochs and the loss is minimized with respect to all network parameters except for $T$. Then a second stage of tuning is entered where we minimize the loss with respect to the parameter $T$ and the validation set. We repeat this process for several thousand training epochs until the performance on the validation set no longer improves.

\begin{figure}[h]
{
%{\includegraphics[scale=0.8]{binToriginal.png}}
%	\includegraphics[scale=0.8]{bininT0.png}
%  {\includegraphics[scale=0.83]{bininT1.png}}
%  {\includegraphics[scale=0.80]{bininT2.png}}
  
%  \includegraphics[scale=0.80]{binToriginal.png}
 % 	\includegraphics[scale=0.84]{2lbpT0.png}}
 % {\includegraphics[scale=0.84]{2lbpT0_05.png}}
%  {\includegraphics[scale=0.844]{2lbpT0_1.png}
  
  }
  %\hspace{115pt}
  
  {
{\includegraphics[scale=0.8]{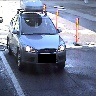}}
	\includegraphics[scale=0.8]{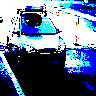}
  {\includegraphics[scale=0.8]{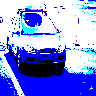}}
  {\includegraphics[scale=0.8]{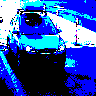}}
  
  \includegraphics[scale=0.8]{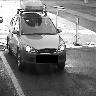}
  	\includegraphics[scale=0.842]{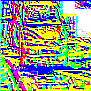}}
  {\includegraphics[scale=0.845]{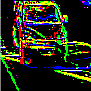}}
  {\includegraphics[scale=0.852]{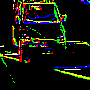}}

  \caption{Input binarization with RGB Thresholding (first row) and LBP (second row).}
\end{figure}
  \label{fig:binarizedInputs}

\textbf{Local Binary Patterns (LBP)}
A well-known technique called \textit{local binary patterns} for extracting multi-resolution and scale-invariant features from images has been introduced in [\ref{refLBP}]. We use a similar approach in our application for image binarization, but with a slight modification: we operate on the grayscale image and process each pixel by examining its neighborhood at a radius of 1 pixel, generate 3 artificial color channels and select 3 pixels at a clockwise stride of $3$ in the neighbourhood to distribute to these channels. Then the value of these pixels are set to $1$ if they exceed the value of the center pixel and $0$ otherwise. An example of this transformation on an image from the dataset is demonstrated in the second row of Figure \ref{fig:binarizedInputs}.

%\jani{Here it would be appropriate to add a short section 2.4 that would explain packing of binarized weights. In the present text format, packing suddenly is mentioned in 3.1 without explanation.}
%\response{I'm not sure if that would be necessary considering that it's pretty simple and the lack of space, but I will try to squeeze it in here. I wrote a short section below, but I feel like it needlessly complicates things.}
\subsection{Packing binary-valued vectors}
To avoid confusion with terminology, we denote by \textit{packing} the encapsulation/conversion of an array of 1-bit values into an individual 32-bit unsigned integer. Formally, for a binary-valued vector $\mathbf{x}\in \left\lbrace-1,+1\right\rbrace ^{D}$, assuming $D$ is divisible by $B$, then the packed representation of $\mathbf{x},$ $\mathbf{x}_p\in \left\lbrace-1,+1\right\rbrace ^{D/B}$
%$\mathbf{x}_p\in \left\lbrace 0,1,2,3,...,2^{B-1}\right\rbrace ^{D/B}$ 
for a packing bitwidth $B\leq32$ (assuming 32-bit word) and positive $D$, is given by 
% \begin{equation}
% \mathbf{x}_p= \begin{bmatrix}
%     \sum^{B-1}_{i=0}\frac{(1+x_i)(B-1 - i)}{2}\\[0.3em]
%      \sum^{2B-1}_{i=B}\frac{(1+x_i)(B-1 - i)}{2}\\[0.3em]
%        \vdots \\[0.3em]
%  \sum^{D}_{i=D-B}\frac{(1+x_i)(B-1 - i)}{2}\\[0.3em]
%      \end{bmatrix},
% \end{equation}

% \begin{equation}
% \mathbf{x}_p= \begin{bmatrix}
%     \sum^{B-1}_{i=0}(1+x_i)2^{B-2 - mod(i,B)}\\[0.3em]
%      \sum^{2B-1}_{i=B}(1+x_i)2^{B-2 - mod(i,B)}\\[0.3em]
%        \vdots \\[0.3em]
%  \sum^{D}_{i=D-B}(1+x_i)2^{B-2 - mod(i,B)}\\[0.3em]
%  \sum^{D}_{i=D-B}(1+x_i)2^{B-2 - mod(i,B)}\\[0.3em]
%      \end{bmatrix}.
% \end{equation}

\begin{equation}
\mathbf{x}_p= \begin{bmatrix}
    \sum^{B}_{i=1}(1+x_i)2^{B-2 - mod(i-1,B)}\\[0.3em]
     \sum^{2B}_{i=B+1}(1+x_i)2^{B-2 - mod(i-1,B)}\\[0.3em]
     \sum^{3B}_{i=2B+1}(1+x_i)2^{B-2 - mod(i-1,B)}\\[0.3em]
       \vdots \\[0.3em]
 %\sum^{D}_{i=D-B}(1+x_i)2^{B-2 - mod(i-1,B)}\\[0.3em]
 \sum^{D}_{i=D-B+1}(1+x_i)2^{B-2 - mod(i-1,B)}\\[0.3em]
     \end{bmatrix}.
\end{equation}

% \begin{equation}
% \mathbf{x}_p= \frac{1}{2}\begin{bmatrix}
%     \sum^{B}_{i=1}(1+x_i)2^{B - i}\\[0.3em]
%      \sum^{2B}_{i=B+1}(1+x_i)2^{2B - i}\\[0.3em]
%      \sum^{3B}_{i=2B+1}(1+x_i)2^{3B - i}\\[0.3em]
%        \vdots \\[0.3em]
%  %\sum^{D}_{i=D-B}(1+x_i)2^{B-2 - mod(i-1,B)}\\[0.3em]
%  \sum^{D}_{i=D-B+1}(1+x_i)2^{D - i}\\[0.3em]
%      \end{bmatrix}.
% \end{equation}

\section{Implementation}
%\section{Efficient Software Implementation of a Binarized Convolutional Network}

In this section, we present the details of our CUDA implementation of the binarized neural network architecture described in Section 2. We use CUDA terminology throughout this section.

\subsection{Convolutional layers}

% Implementations:
% cuDNN fp
% cuDNN + bin
% cuDNN + bin with bin inputs
% OpenCL baseline
% ARM CL

% On NVIDIA platforms we compare our implementation with the fp cuDNN implementation. 
% We implement the binarized neural network in OpenCL and CUDA and benchmark our implementation on three different GPUs. 

%\jani{one would expect that this section would start by an overview of the binarization approach, however now it starts by GEMM}
%\response{: yes, this section was initially called "Convolutional layers", but then Heikki suggested this change because the last section was short, but I think we can organize them into convolutional layers and matrix multiplications (includes second stage in convolution + dense layers)}

The convolutional layer in a neural network can significantly improve image classification accuracy compared to standard multi-layer perceptrons. Given a kernel $H\in \mathbb{R} ^{K \times K\times C  }$ and an image $ X\in \mathbb{R}^{ H\times W \times C}$, an output feature map $ F \in \mathbb{R}^{ H\times W} $ is given by the expression
% \expendable{4}
% \prosaic{
% \begin{equation}
% F[m,i,j]=\sum^{C=1}_{c=0} \sum^{K-1}_{l=0} \sum^{K-1}_{k=0} H[m,l,k,c]X[i-,j-.,c]...
% \end{equation}
% }

\vspace{-2.8mm}
\begin{equation}\label{xnormult}
    %\resizebox{1.20\hsize}{!}{%
         \hspace{-1.2mm} F[i,j]=\hspace{-1.0mm}
         \sum^{C-1}_{c=0} \sum^{R}_{l=-R} \sum^{R}_{k=-R} \hspace{-1.0mm} H[R+l,R+k,c]X[i+k,j+l,c],\hspace{-1.5mm}  
     %   }
\end{equation}
for odd $K$, and the kernel radius \resizebox{0.15\hsize}{!}{$R=\frac{K-1}{2}$}.
It should be noted that equation (3) in fact computes cross-correlation (not convolution), which is the convention in deep learning. A common approach for computing convolutions efficiently is through matrix multiplication [\ref{refGEMMConv}], where the weights and image tensors are reshaped into 2-dimensional matrices, which will then allow us to compute the convolution through a single matrix multiplication. The reshaping for the weights is trivial, and this step can often be skipped if the weights are already stored in this layout; however, the process of arranging the input image into the matrix of columns used for computing the convolution can be difficult to optimize. This is due to inefficient access patterns, complicated index calculations that involves many division and modulo operations, and the overhead of storing the large output matrix to global memory.
%\jani{I tried to re-formulate the latter half of the baragraph just above, please check if it makes sense}
%\jani{As you had already planned, an explanation about im2col would be good here. You mention this function 3 paragraphs from here.} \response{I was thinking may I should just explain im2col implicitly as done here and never refer to it as 'im2col' because it sounds a little informal.}
%\jani{It's fine if you omit the description of im2col if there is no reference to it later.}
%In our case, the latter problem 
%\prosaic{The function that does this is often called \textit{im2col3d} , and we will refer to it like that from here on.}

A straightforward approach for avoiding inefficient access patterns is to load regions from the image into shared memory (on-chip memory) and then extract the patches from shared memory [\ref{refCUDNN}]. For an image with dimensions $H\times W\times C$ corresponding to height, width, and channels respectively, and a $K\times K \times C$ kernel with a radius of $R=\frac{K-1}{2}$, we use threadblock dimensions of $S\times W$ ($S=2$ in our case), which covers the entire width of the image, eliminating the need to redundantly load the horizontal non-zero halo regions which are difficult to load with an efficient access pattern. Then each thread-block loads an image region of dimensions $(S + 2R)\times W$ into a region in shared memory in three steps, starting by loading the top vertical halo region, the middle part, then the bottom vertical halo region (except when loading from the bottom of the image). The shared memory buffer is zero-initialized in order to implicitly handle horizontal zero-padding. Loading vertical halo regions can be done very efficiently since all threads in the threadblock load from contiguous regions in the image array.
%The allocated region in shared memory is pre-initialized to zero, and padding is simply handled by shifting the loaded image block in the destination in shared memory.

In the second stage, the patches of size $K\times K \times C$ are extracted from shared memory.  We avoid division and modulo operations in the patch-extraction stage by using an integer counter register. This results in a 2$\times$ performance boost in our case. Since the network is binarized, the packing and patch-extraction step can be fused into one step to avoid redundant accesses to global memory, reducing global memory stores by $K\times K$. The algorithm for the combined step of extracting the patches and packing them is shown in Algorithm 1.

%\prosaic{Even the full-precision version of our \texttt{im2col} algorithm, which does not involve output binarization, is 2$\times$ faster than in cuDNN, which we believe is due to eliminating slow division and modulo operations.}

% A major bottle-neck in the im2col algorithm can [is the part where] the outputs are stored into global memory. The input to the algorithm is a HxWxC image, however, the output can have a KxK fold increase in size. In a binarized neural network however, this problem can be avoided by only storing the packed representation of the output matrix to global memory. Each thread handles a $K\times K$ segment in the image , but since our kernel outputs a packed columns matrix, each thread stores a single 32-bit unsigned integer register in the columns matrix and in a fully coalesced manner, where each register contains the packed representation of all $K\times K$ pixels in the image patch, reducing the amount of global memory stores by 25X in our case for a kernel size of $5\times 5$. 
%This process is illustrated in Figure \ref{fig:figIm2col}.

% \begin{figure}[h]
% \center  {\includegraphics[scale=0.2]
%   {bim2col3d.png}}
 
%   \caption{\expendable{1} DO A BETTER VERSION LATER}  
%     \label{fig:figIm2col}
% \end{figure}

\begin{algorithm}
\caption{Patch-extraction and packing}\label{}
\begin{algorithmic}[1]
\Function{\textup{ ExtractPacked}}{sh\_block}:
%\State	 $s \gets (W + 2 \times R) \times \texttt{\_t}_y+\texttt{\_t}_x$
\State  $v \gets 0$
\State $k  \gets 0$
\For {$i = 0$ to $B-1$}

\If {($i - k K =K$)}
\State			$k++$
\EndIf
%\State		$idx$ = $s + k \times (W + 2\times R) + i - k \times K$
\State		$idx =(W+2R)(\texttt{\_t}_y + k) +\texttt{\_t}_x + i -kK$
\State		$s =$ sh\_block[$idx$]$\hspace{1.5mm}>\hspace{1.5mm}0$

\State		$v$ = bitOR( $v, s << B - 1 - i$)
\EndFor

\State \textbf{return $v$}
\EndFunction
\end{algorithmic}
\end{algorithm}

In Algorithm 1, \texttt{sh\_block} is the region of the image loaded into shared memory using the previously described steps, including the halo regions. $\texttt{\_t}_x$ and $\texttt{\_t}_y$ are the thread indices for the $x$ and $y$ dimensions of the thread block corresponding to the CUDA \texttt{threadIdx.x} and \texttt{threadIdx.y} variables. $B$ is the packing bitwidth, chosen to be $25$ in our case, $<<$ is the left bit-shift operator, and $v$ is the packed extracted patch.

%\texttt{\textbf{comment:} section names 3.1 and 3.2 are not very logical when compared to each other. If section 3.1 is about convolution layers, it would be logical that 3.2 would concern dense layers} \response{I tried to change it, but now the dense layer section is kinda short.}

For computing the convolution we implement a standard matrix multiplication subroutine in a manner similar to [\ref{refDGEMM}], where tiles from each matrix are loaded successively into shared memory and used to compute a submatrix of the output, such that each thread computes a single element in the output matrix, but instead of computing multiplications, we compute xnors and bit-counts following an approach similar to what was suggested in [\ref{refBinNet}] as
\vspace{-1.2mm}
\begin{equation}\label{xnormult}
\vspace{-1.2mm}
         \mathbf{a\cdot}\mathbf{b}= \texttt{W}- \texttt{2}\times \texttt{popcount(xor(}A,B\texttt{))},% 
\end{equation}
where ${A}$ and ${B}$ are both 32-bit unsigned integer registers containing the packed representations of vectors \textbf{a}, \textbf{b} $\in \left\lbrace-1,+1\right\rbrace ^{\texttt{W}}$ respectively. We denote by $\cdot$ the real-valued dot product. The operation \texttt{xor} is the bit-wise xor operation, and \texttt{popcount} is a function for computing the number of bits set to $1$. The packing bitwidth \texttt{W} is the number of elements that are packed together in a single unsigned integer register. %\prosaic{The output matrix is in full-precision and identical in dimensions to the full precision version. The results are also identical as long as the matrices only take on the values $-1$ and $+1$. }

\subsection{Fully connected layer}
For the fully-connected layer, we follow a slightly different approach from standard matrix multiplication. For a packed weights matrix $\bf{W}\in \mathbb{R}^{L\times D}$, and a packed vector $\bf{x}\in \mathbb{R}^{D\times 1}$, we divide the process of computing the dot product of each weight vector and $\bf{x}$ into $64$ segments, such that each of 64 threads handling a weight vector compute the partial sum of the dot product between a weight vector and $\bf{x}$ through xnor operations, and stores the results in shared memory. The partial sums are then combined in a parallel reduction sum that does not require synchronization (for a warp size of 32 on the target platform).
%\jani{is this on the left 64 times L or 64 times x times L? The possibility of confusion is great as there is the vector of name x involved}
\section{Results}

 In this section we present our results for the impact of input binarization on classification accuracy and the performance improvement achieved.

% In this section we present the results [of performane boost] from our implementation compared to an equivalent full-precision implementation on the target platforms. [for how] input binarization impacts a binarized neural network in terms of runtime speed, storage requirement, and classification accuracy. Additionally we [present] the results of our implementation and compare it with existing full-precision implementations that [use] can achieve state of the art performance on each platform we test on and 

\textbf{Input binarization}
in Table \ref{tblAccuracy} we report the classification accuracy results we obtained using each different input binarization scheme for our binarized version of the vehicle classifier [\ref{refHeikkiCar}]. We can observe that accuracy is best retained when the first layer is not binarized; however, only a moderate loss in accuracy occurs when using LBP and RGB Thresholding. Considering that RGB Thresholding is much simpler to implement and results in almost no additional computational overhead, we choose this approach for our final binarized architecture, for which we report the speed up results in the following section.
% (accuracy table)
%  In TABLE2 we report the best result we have obtained using this method with a threshold value of $0.01$ for pixels in range $0.0 - 1.0$.

\begin{table} 
\caption{Runtime of the network on each platform}
\begin{center} % Centers the table on the page, comment out to left-justify
\resizebox{\columnwidth}{!}{%
\begin{tabular}{l c c c c} % The final bracket specifies the number of columns in the table along with left and right borders which are specified using vertical bars (|); each column can be left, right or center-justified using l, r or c. To specify a precise width, use p{width}, e.g. p{5cm}
%\toprule % Top horizontal line
& \multicolumn{3}{c}{} \\ % Amalgamating several columns into one cell is done using the \multicolumn command as seen on this line
%\cmidrule(l){2-4} % Horizontal line spanning less than the full width of the table - you can add (r) or (l) just before the opening curly bracket to shorten the rule on the left or right side
% Joint & LSTM1 & LSTM2 & LSTM3 \\ % Column names row
% \midrule % In-table horizontal line
% MAE		&0.0196 &  0.0167 & 0.0188 \\ 
% MID  &	0.2034& 0.2185 & 0.1437 \\  
% Energy$_{avg}$ & 2.6512 & 2.8149& 1.9892\\
% Energy$_{std}$ & 1.2017 & 1.9342 & 1.0882\\
% \midrule % In-table horizontal line
% \bottomrule % Bottom horizontal line

Implementation Method & GTX1080 & Mali T860 & Tegra X2 \\ % Column names row
\midrule % In-table horizontal line
%baseline (full-precision)  & 0.0 $\mu s$  & 0.0 $\mu s$ & {0.0 $\mu s$} \\ 
cuDNN (full-precision)	& $401.83 \mu s$  & N/A$^\dag$  &  $2.27$  ms \\ 
Arm CL (full-precision)	& N/A$^\dag$   & $29.61$ ms  &  N/A$^\dag$ \\ 
BCNN   & $102.39 \mu s$ &  $23.63$ ms & $0.53$ ms \\
BCNN with binarized inputs  & $\mathbf{55.63} \mu s$ & $\mathbf{17.58}$ ms & $\mathbf{0.41}$ ms\\

\midrule % In-table horizontal line
\bottomrule % Bottom horizontal line

\end{tabular}
}
\end{center}
 $^\dag$Library not compatible with this platform.
 % Table caption, can be commented out if no caption is required
\label{tblTime} % A label for referencing this table elsewhere,

\end{table}

\begin{table} 

\caption{Runtime per-layer (GTX1080)}
\centering % Centers the table on the page, comment out to left-justify
\resizebox{\columnwidth}{!}{
\begin{tabular}{l c c c c} % The final bracket specifies the number of columns in the table along with left and right borders which are specified using vertical bars (|); each column can be left, right or center-justified using l, r or c. To specify a precise width, use p{width}, e.g. p{5cm}

\toprule % Top horizontal line
& \multicolumn{3}{c}{} \\

% Amalgamating several columns into one cell is done using the \multicolumn command as seen on this line
 % Horizontal line spanning less than the full width of the table - you can add (r) or (l) just before the opening curly bracket to shorten the rule on the left or right side
% Joint & LSTM1 & LSTM2 & LSTM3 \\ % Column names row
% \midrule % In-table horizontal line
% MAE		&0.0196 &  0.0167 & 0.0188 \\ 
% MID  &	0.2034& 0.2185 & 0.1437 \\  
% Energy$_{avg}$ & 2.6512 & 2.8149& 1.9892\\
% Energy$_{std}$ & 1.2017 & 1.9342 & 1.0882\\
% \midrule % In-table horizontal line
% \bottomrule % Bottom horizontal line

Layer & cuDNN & Binarized & Speed-up \\ % Column names row
\midrule % In-table horizontal line
%baseline (full-precision)  & 0.0 $\mu s$  & 0.0 $\mu s$ & {0.0 $\mu s$} \\ 
Im2col3d $(96,96,3)$	& $21.63$ $\mu s$  & $3.17\mu s$ & $6.82\times$ \\ 
GEMM-convolution $(32,5,5,3)$	& $37.54 \mu s$   & $8.61\mu s$ &  $4.36\times$  \\ 
Max-Pooling $(96,96,32)$	& $5.22 \mu s$ & $8.26 \mu s$   & $0.63\times$\\ 
Im2col3d $(48,48,32) $  & $65.41 \mu s$ &  $5.50 \mu s$ &  $11.89\times$\\
GEMM-convolution $ (32,5,5,32)$	  & $69.28 \mu s$ & $8.10 \mu s$ &  $8.55\times$\\
Max-Pooling (48,48,32)	& $5.38 \mu s$    & $2.66 \mu s$  &  $2.02\times$\\ 
Fully-Connected $(100, 24\times 24 \times32)$  & $200.03 \mu s$ & $6.28 \mu s$ &  $31.85\times$\\

\midrule % In-table horizontal line
\bottomrule % Bottom horizontal line

\end{tabular}
}
 % Table caption, can be commented out if no caption is required
\label{tblSpeed} % A label for referencing this table elsewhere,

\end{table}

\textbf{Performance Boost}
We time our binarized implementation on 3 different hardware platforms: Nvidia GTX 1080, Nvidia Jetson (Tegra X2), and the Mali-T860. We derive an OpenCL version of our implementation for testing on the Mali-T860, which is a straightforward process. We compare the performance of our implementation against an equivalent full precision version of the same network implemented with highly optimized libraries on each target platform, in our case these are cuDNN on Nvidia platforms, and the ARM Compute Library on the Mali-T860. 
We list in Table \ref{tblTime} the average execution times of the full network on each platform. \prosaic{We can see that} our binarized implementation can achieve up to $7.5\times $ speed up on the GTX1080 and about $5.5\times $ on the Tegra X2. \prosaic{We also notice that} the relative performance improvement on Mali GPU is much smaller at about $1.7\times $ for the fully binarized version. In our optimizations, we heavily take advantage of using local memory (in OpenCL terms) which resides on-chip in most workstation GPUs and the Nvidia Tegra X2, but this does not offer any performance benefits on Mali GPUs since local memory is allocated in global memory.
It should be noted that cuDNN is optimized for batch processing and that our results are for one sample at a time which means these results may not necessarily be reflective of the full potential of cuDNN; however, batch processing is not a suitable option for real-time applications where a single input is processed at a time. Additionally, we note that for our cuDNN implementations, we use the explicit GEMM convolution algorithm, which can be slightly slower than the implicit GEMM algorithm. For example, cuDNN with implicit GEMM can run at $316 \mu s$ for the first convolutional layer in our network on the GTX1080.%, and at $1.626$ ms on the Tegra X2.

For a more detailed comparison, we present the execution times for each individual layer in Table \ref{tblSpeed}. Each layer's name is followed with the dimensions of the input, except for the convolution layers where the dimensions are for the kernels, and the input dimensions can be inferred from the previous layer. This table compares the execution time of our binarized implementations with the full-precision versions of the same layer in cuDNN on the GTX1080. We omit from the table the computation times for ReLU activations, which are present in the full-precision version of the network, but are absent from the binarized version. We also omit the last 2 fully-connected layers since they are too small and in most practical applications it would be more efficient to implement them on the CPU. We include the computation time for packing the outputs of the previous layer in the binarized version of the fully-connected layer for a fair comparison. The results in Table \ref{tblSpeed} have been obtained directly from the Nvidia Visual Profiler. %after the results of the first fully-connected layer has been obtained. 

It should be noted that the runtime for the fully-connected layer for full-precision cuDNN in Table \ref{tblSpeed} includes a matrix transposition. The run time excluding matrix transposition is about $100 \mu s$; however, it is a necessary step for evaluating this layer. Our full-precision matrix multiplication kernel is in fact $2\times$ slower than cuBLAS (as measured in this network), yet a significant speed-up is still achievable through binarization.

\begin{table} 
\caption{Impact of different input-binarization schemes on classification accuracy}
\centering 
%\resizebox{\columnwidth}{!}{%
\begin{tabular}{l c c c c} 
& \multicolumn{1}{c}{} \\ 

Method & Accuracy \\ % Column names row
\midrule % In-table horizontal line
LBP	&  {92.06\%}\\ 
Thresholding  Grayscale& ${89.16\%}$ \\
Thresholding RGB &  ${92.52\%}$ \\
No input binarization    &${94.20\%}$ \\ 
Full-precision network    &${97.09\%}$ \\ 
\midrule % In-table horizontal line
\bottomrule % Bottom horizontal line

\end{tabular}
%}
% Table caption, can be commented out if no caption is required
\label{tblAccuracy} % A label for referencing this table elsewhere,
%\response{not sure if we should include the multiplications and additions. They feel like they were forced into the table just to fill space. But if needed, I can still add that part.}
%\jani{Yeah, I think you can omit the muls and adds. Time is anyhow running short, so it is better to concentrate on polishing the paper otherwise. If needed, those can be added for the camera ready if the paper is accepted.}
\end{table}

\section{Conclusion and Future Work}
We presented an efficient implementation of a binarized convolutional neural network on GPUs that can achieve a significant decrease in runtime while reasonably preserving classification accuracy. In the future we wish to restructure our algorithms to achieve a similar performance improvement on other embedded platforms. We are also planning to extend this work to alternative convolution algorithms such as implicit GEMM, which can be faster than explicit GEMM. Finally, we plan to extend our study of how input binarization impacts classification accuracy on larger datasets with more difficult classification tasks.
% {We also wish to further the work on vehicle type classification. Vheicle type classification can be difficult, for one, it is difficult to assign a single class. In future, option is to use multi-label classification. Another alternative approach is to evaluate accuracy through top-2 or top-3; however, this would require a larger dataset and a narrower definition of vehicle types and it may not be fair to use the top-2 or top-3 accuracy in our case since th our dataset is categorized into 4 classes.} 


% \section{Supplementary Material}
% Video links showing the motion sequences generated by each model. Sequences with the blue skeleton are the input sequences feed to the network. Motion sequences in green are generated by the network.
% \begin{itemize}
% \item \href{https://www.youtube.com/watch?v=v_WTs5EXT3c}{LSTM1}
% \item  \href{https://www.youtube.com/watch?v=jTxqGgwFWPk}{LSTM2}
% \item  \href{https://www.youtube.com/watch?v=9-Eol83xZZA}{LSTM3}
% \end{itemize}

%\ifCLASSOPTIONcompsoc
%This work was supported supporters.
%\else
\section*{Acknowledgment}
This work was funded by the Academy of Finland
project 309903 CoEfNet.

%\vspace{-4mm}

%\fi



%\bibliographystyle{IEEEtran}

\end{document}